\documentclass{article}
\usepackage{ijcai21}

\usepackage{times}
\usepackage{soul}
\usepackage{url}
\usepackage[hidelinks]{hyperref}
\usepackage[utf8]{inputenc}
\usepackage[small]{caption}
\usepackage{graphicx}
\usepackage{amsmath}
\usepackage{amsthm}
\usepackage{booktabs}
\usepackage{algorithm}
\usepackage{algorithmic}
\usepackage{amsfonts,amssymb}
\usepackage{comment}

\usepackage{epsfig,epstopdf}
\usepackage{subfigure}
\usepackage{color}
\usepackage{multirow}

\title{Cross-modal Zero-shot Hashing by Label Attributes Embedding}

\author{
Runmin Wang$^{1,2}$\and
Guoxian Yu$^{2,3,4}$\footnote{Contact Author}\and
Lei Liu$^{2,3}$\And
Lizhen Cui$^{2,3}$\and
Carlotta Domeniconi$^{5}$\and
Xiangliang Zhang$^{4}$\\
\affiliations
$^1$College of Computer and Information Sciences, Southwest University, Chongqing, China.\\
$^2$School of Software, Shandong University, Jinan, China.\\
$^3$Joint SDU-NTU Centre for Artificial Intelligence Research, Shandong University, Jinan, China.\\
$^4$CEMSE, King Abdullah University of Science and Technology, Thuwal, SA.\\
$^5$Department of Computer Science, George Mason University, China.\\
\emails rmwang@email.swu.edu.cn, \{gxyu, l.liu, clz\}@sdu.edu.cn, carlotta@cs.gmu.edu, xiangliang.zhang@kaust.edu.sa
}

\begin{document}

\maketitle

\begin{abstract}
Cross-modal hashing (CMH) is one of the most promising methods in cross-modal approximate nearest neighbor search.
Most CMH solutions ideally assume the labels of training and testing set are identical. However, the assumption is often violated, causing a zero-shot CMH problem. 
Recent efforts to address this issue focus on
transferring knowledge from the seen classes to the unseen ones using label attributes. However, the attributes are  isolated from the  features of multi-modal data. 
To reduce the information gap, we introduce an approach called LAEH (Label Attributes Embedding for zero-shot cross-modal Hashing). LAEH first gets the initial semantic attribute vectors of labels by word2vec model and then uses a transformation network to transform them into a common subspace. Next, it leverages the hash vectors and the feature similarity matrix to guide the feature extraction network of different modalities. At the same time, LAEH uses the attribute similarity as the supplement of label similarity to rectify the label embedding and common subspace. Experiments show that LAEH outperforms related representative zero-shot and cross-modal hashing methods.
\end{abstract}

\section{Introduction}

Approximate nearest neighbor (ANN) search has been playing an increasing important role in many applications, e.g., image retrieval in a large database \cite{wang2014hashing,wang2017L2Hsurvey} Hashing is a kind of technique that transforms the data points from the original feature space into a compact Hamming space with binary hash codes, which can   efficiently  solve the problem of approximate nearest neighbor search. Recently, hashing has attracted much research interest because it can substantially reduce the storage cost and dramatically improve the query speed.
In particular, hashing methods have been designed for the multi-modal situation, e.g.,  finding a picture or a video with a keyword.  
These cross-modal hashing (CMH) methods  ~\cite{abelson-et-al:scheme,wang2017L2Hsurvey} aim to find hash function for different modalities respectively. Existing CMH methods can be roughly divided into two categories: supervised and unsupervised. Supervised CMH methods utilize the supervising information (semantic labels or similarity relationship) to guide the exploration of intrinsic data property, while unsupervised methods only take advantages of the data internal structure. Generally, supervised methods perform better than unsupervised methods.


A common assumption made in supervised CMH methods is that labels of the training and testing set are identical. However, this assumption is often violated in real-world applications, causing zero-shot CMH problem. In order to work on ``unseen'' categories, most current zero-shot CMH methods introduce extra attribute information for each class label or each sample. They learn hash functions on attribute space ~\cite{ji2020attribute-guided} or transfer attribute knowledge from seen classes into unseen classes ~\cite{liu2019cross,xu2020ternary}. The major limitation of these methods is that {the label attribute information is not compatible with the multi-modal data information, as they come from different sources.  }

In this paper, we propose a novel cross-modal hashing method called Label Attribute Embedding based cross-modal Hashing (LAEH). Our main contributions are   as follows:
\begin{itemize}
\item LAEH   introduces a label  embedding network, which addresses the incompatibility  between the label attribute information and the   multi-modal data information. The multi-modal features and label embedding features are mapped to a common space, facilitating the measurement of their similarity.


\item LAEH leverages the attribute similarity derived from embedded  label attributes and the similarity induced from multi-modal data, resulting an adaptive model fitting to the zero-shot learning paradigm.
\item We perform experiments on two benchmark datasets and the results show that LAEH achieves state-of-the-art performance in zero-shot CMH. 
\end{itemize}

\section{Related Work}
In general, CMH methods can be divided into two main categories: supervised and unsupervised, depending on the use of semantic labels (class labels) of instances. We focus the discussion here on the supervised CMH methods, as our study takes labels into usage as well. 


Supervised methods additionally use the label information of training data, which contains important semantic information, and often achieve a better performance than unsupervised ones. To name a few, CMSSH ~\cite{bronstein2010data} regards every bit of hash code as a classification task and learns the whole hash code bits one by one. SCM ~\cite{zhang2014large-scale} optimizes the hashing functions by maximizing the correlation between two modalities with respect to the semantic similarity obtained from labels.
SePH ~\cite{lin2015semantics} transforms the semantic affinities obtained from class labels into a probability distribution and approximates it with to-be-learned hash codes in the Hamming space, and then learns the hash codes via kernel logistic regression. CRE ~\cite{hu2019collective} processes heterogeneous types of data using different modality-specific models and unifies the projections of different modalities to the Hamming space into a common reconstructive embedding to obtain hashing codes. WCHash ~\cite{liu2019weakly} firstly optimizes a latent central modality with respect to other modalities. Then it enriches the labels of training data with features to deal with the incomplete supervised information. Finally it uses the enriched similarity to guide correlation maximization between the respective data modalities and the central modal to overcome multi-molal problem and weakly-supervised problem simultaneously. FlexCMH ~\cite{yu2020flexible} introduces a clustering-based matching strategy to explore the structure of each cluster to find the congruent relationship of these clusters among different modalities to handle the weakly-paired multi-modal data.

Recently, CMH methods with deep architectures have been extensively studied, e.g., DCMH ~\cite{jiang2017deep}, SSAH ~\cite{li2018self-supervised}, and CYC-DGH ~\cite{wu2019cycle-consistent}. These deep methods take advantage of the strong feature extraction ability of deep neural networks and often achieve a significant performance. RDCMH ~\cite{liu2019ranking} first calculates the semi-supervised semantic similarity and derives a similarity ranking list, then it optimizes the feature learning model and hash learning model simultaneously with the triplet ranking loss.
ASCSH ~\cite{meng2020asymmetric} decomposes the hash mapping matrices into a consistent and a modality-specific matrix to sufficiently exploit the intrinsic correlation between different modalities, and then uses a novel discrete asymmetric framework to explore the supervised information and solves the binary constraint problem without any relaxation.

With the emergence of a large number of new concepts, the zero-shot CMH problem has received more attention, and many zero-shot CMH methods have been proposed to  deal with emerging (unseen) classes in CMH problem.
Attribute Hashing (AH)~\cite{xu2017attribute},  a single modal zero-shot hashing method,  introduces a multi-layer matrix factorization to exploit attributes and to model the relationships among visual features, binary codes and labels. AgNet ~\cite{ji2020attribute-guided} aligns different modal data into a semantically rich attribute space, and then transforms attribute vector into hash codes via a hash code generation network. TANSS ~\cite{xu2020ternary} adopts an adversarial network to coordinate deep hash functions learning across modalities and knowledge transfer from seen to unseen classes. CZHash ~\cite{liu2019cross} first quantifies the composite similarity between instances using label and feature information, and then induces hash functions by composite similarity preserving and category attribute space learning.


\section{Proposed LAEH Method}


\subsection{Notations and Problem Formulation}

We use boldface uppercase letters (like $\mathbf{F}$) to represent matrices, and boldface lowercase letters (like $\mathbf{v}$, $\mathbf{f}$) to represent vectors. The same letters represent the same matrix and its row (column) vector without special instructions. For example, $s_{ij}$ represents the element in the $i$-th row and $j$-th column of matrix $\mathbf{S}$, $\mathbf{F}_{*j}$ represents the $j$-th column of $\mathbf{F}$, and $\mathbf{F}_{i*}$ represents the $i$-th row of $\mathbf{F}$. $\mathbf{F}^T$ is the transpose of $\mathbf{F}$. We use $tr(\cdot)$ and $\parallel\cdot\parallel_F$ to denote the trace and Frobenius norm of a matrix, and $sgn(\cdot)$ is the signum function.

Our training set $\mathcal{O}_{train}=\{o_i\}_{i=1}^n$ includes instance $o_i=\{\mathbf{x}_i, \mathbf{y}_i\}$  with two modalities, where $\mathbf{x}_i$ is the image feature vector, and $\mathbf{y}_i$ is the textual feature vector of the $i$-th instance. Matrix $\mathbf{X}^{(1)}$ and $\mathbf{X}^{(2)}$ denote the data matrix of modal $\mathbf{x}$ and modal $\mathbf{y}$, respectively. Matrix $\mathbf{L}\in\{0,1\}^{n\times l_s}$ represents labels, where $l_{ij}=1$ indicates that the $i$-th instance belongs to the $j$-th category and $l_{ij}=0$ indicates that the $i$-th instance does not belong to the $j$-th category. The labels of all instances in the training set are ``seen'' labels, collected in set $\mathcal{C}^{(s)}$ which contains $l_s$ distinct category labels. $\mathbf{S}\in\{0,1\}^{n\times n}$ is the similarity matrix, where $s_{ij}=1$ means instance $i$ and $j$ are similar to each other (belong to the same category) and $s_{ij}=0$ means instance $i$ and $j$ are not similar (belong to different categories).
The goal of zero-shot cross-modal hashing is to learn hash codes for each instance and to learn different hash functions $H_1(\mathbf{x}_i):\mathbb{R}^{d_1}\rightarrow \{-1,1\}^c$ and $H_2(\mathbf{y}_i):\mathbb{R}^{d_2}\rightarrow \{-1,1\}^c$, projecting features learned from different modalities to a low dimensional Hamming space, respectively. Note that in zero-shot learning paradigm, the labels of instances in the test set $\mathcal{C}^{(t)}$ may belong to the ``seen'' category set $\mathcal{C}^{(s)}$ or the ``unseen'' category set $\mathcal{C}^{(u)}$, where $\mathcal{C}^{(s)}\cap\mathcal{C}^{(u)}=\emptyset$ and $\mathcal{C}^{(u)}$ contains $l_u$ different category labels.

\begin{figure} [tb]
  \centering
  \includegraphics[width=8.5cm, height=4cm]{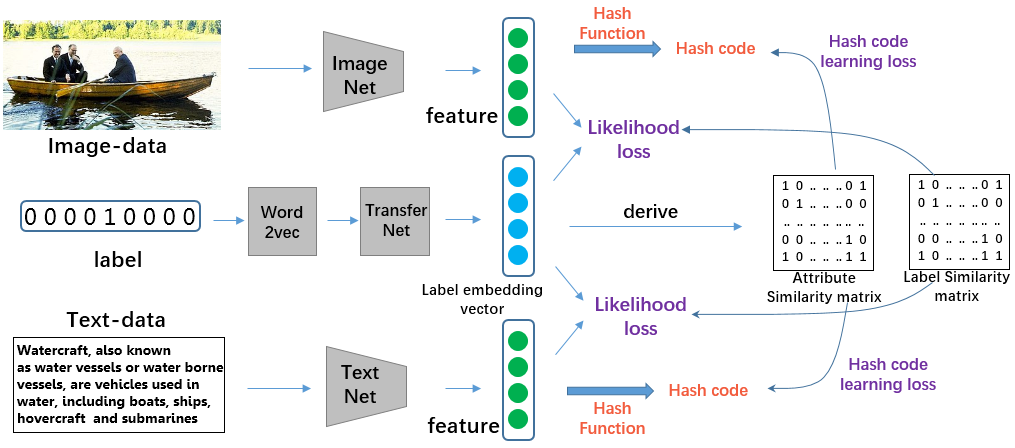}
  \caption{The  architecture of the proposed label attribute embedding hashing (LAEH) method. LAEH first obtains semantic vector for each sample and then uses deep neural networks to learn deep features and label embedding vectors. Then it derives attribute similarity matrix from label embedding vector and uses the attribute similarity as a supplement for label similarity to guide the hash code learning. Finally it jointly uses likelihood loss and hash code learning loss to preserve the label similarity and attribute similarity.}
  \label{model}
\end{figure}

To implement zero-shot cross-modal hashing, the main goals of our method are as follows: 1) take advantage of label semantic embedding, learn similar hash codes for instances with similar semantic attributes and make the semantic embedding module compatible with the whole hash code learning framework; and 2) sufficiently exploit the intra-modal correlation as well as inter-modal correlation and learn similar hash codes for instances from the same category. To achieve these goals, we propose a novel label embedding hashing framework, whose   architecture is shown in Fig. \ref{model}.

\subsection{Deep Feature Learning}
Due to the powerful representation learning ability of deep neural networks, we introduce two deep neural networks to extract the features of different modalities. For images, we use a fine-tuned CNN-F ~\cite{chatfield2014return}, which  is originally a deep convolutional neural network with eight layers for image classification. The first five layers are convolutional layers and the last three layers are fully-connected layers. We take the output of the last layer before softmax as the deep features. Note that fine-tuned CNN-F can be replaced by  other deep feature learning models, such as ResNet ~\cite{he2016deep}. For text, we first represent the text into bag-of-words (BOW) vectors, then send them to a neural network with three fully-connected layers and  ReLU  activation functions. 
The learned features are denoted by $\mathbf{F}^{(1)}\in \mathbb{R}^{d\times n}$ and $\mathbf{F}^{(2)}\in \mathbb{R}^{d\times n}$ for image and text, respectively. The parameters of image-net and text-net are denoted as $\theta_x$ and $\theta_y$.

\subsection{Label Semantic Embedding}
To achieve zero-shot learning, attribute information of labels should be leveraged to establish the connection between the seen and unseen labels, for instance, applying a pre-trained  Word2Vec~\cite{mikolov2013distributed} model to get the semantic embedding vectors of labels, noted as $\mathbf{V}\in\mathbb{R}^{v\times n}$.
However, the semantic embedding vectors $\mathbf{V}$ may not be compatible with features $\mathbf{F}^{(1)}$ and $\mathbf{F}^{(2)}$ learned from images and text in the previous step, as the embedding and feature learning were conducted separately. We thus introduce a transform network to transform the semantic embedding vectors $\mathbf{V}$ into label embedding vectors $\mathbf{F}^{(l)}\in \mathbb{R}^{d\times n}$. The transform network is implemented as  a three-layer neural network and the parameters of transform network are denoted as $\theta_l$.  Note that the main goal of our cross-modal hashing is to learn similar hash codes for similar instances. Thus, we expect to learn similar deep feature for similar instances in different modalities. Meanwhile,  the similarity relationships between the label embedding vectors and their deep features in the common subspace should be well preserved. To this end, we introduce the pair-wise semantic embedding loss as:
\begin{equation}
\mathcal{J}_1=\sum_{m=1}^2\sum_{i,j=1}^{n}L([\mathbf{F}^{(m)}_{*i}]^T\mathbf{F}^{(l)}_{*j};s_{ij})
\label{j1}
\end{equation}
where 
$L(\bullet;s_{ij})$ is the semantic embedding loss function, $s_{ij}$ indicates if the $i$-th  and $j$-th  instance have the same label or not ($s_{ij}=1$ if they have, otherwise $s_{ij}=0$). $[\mathbf{F}^{(m)}_{*i}]^T\mathbf{F}^{(l)}_{*j}$ is the inner product between deep feature of modal $m$ of the $i$-th instance and label embedding vector of the $j$-th instance. Obviously, the larger the inner product $[\mathbf{F}^{(m)}_{*i}]^T\mathbf{F}^{(l)}_{*j}$ is, the more similar the deep feature $\mathbf{F}^{(m)}_{*i}$ and label embedding vector $\mathbf{F}^{(l)}_{*j}$ are. Then, when $s_{ij}=1$, the $i$-th instance and the $j$-th instance are  similar, i.e., $\mathbf{F}^{(m)}_{*i}$ and $\mathbf{F}^{(l)}_{*j}$ are similar.
Otherwise, when $s_{ij}=0$, $\mathbf{F}^{(m)}_{*i}$ and $\mathbf{F}^{(l)}_{*j}$ are dissimilar, and $[\mathbf{F}^{(m)}_{*i}]^T\mathbf{F}^{(l)}_{*j}$ is low.

This is analogous to the binary classification problem with label $s_{ij}=1$ or $0$. Let $\Phi_{ij}^m=[\mathbf{F}^{(m)}_{*i}]^T\mathbf{F}^{(l)}_{*j}$.   The loss function $L(\bullet;s_{ij})$ can be defined as
\begin{equation}
\label{likelihood}
\begin{aligned}
L(\Phi_{ij}^m;s_{ij}) &= log(\sigma(\Phi_{ij}^m)^{s_{ij}}(1-\sigma(\Phi_{ij}^m))^{1-s_{ij}}) \\
 &=s_{ij} log(\sigma(\Phi_{ij}^m)) + (1-s_{ij}) log(1-\sigma(\Phi_{ij}^m))
\end{aligned}
\end{equation}
where $\sigma$ is the Sigmoid function.

Taking it into Eq. (1), we have the loss function   $\mathcal{J}_1$ 
\begin{equation}
\mathcal{J}_1=-\sum_{m=1}^2\sum_{i,j=1}^{n}[s_{ij}\Phi_{ij}^m-log(1+\mathrm{e}^{\Phi_{ij}^m})]
\label{j1}
\end{equation}

\subsection{Hash Code Learning}
For zero-shot hash code learning, besides the goal of preserving the similarity w.r.t. category label, we should also keep the semantic relevance w.r.t.  the semantic attributes of instances. Only in this way, we can transfer semantic attribute knowledge from seen labels to unseen labels and achieve zero-shot learning. Thus, we formulate the hash code learning loss as:
\begin{equation}
\mathcal{J}_2=||\frac1c\mathbf{B}^T\mathbf{B}-\mathbf{A}||_F^2
\label{j2}
\end{equation}
where $\mathbf{A}\in \mathbb{R}^{n\times n}$ is the similarity derived from label embedding vectors we obtain in the previous section. Here we use the cosine similarity of label embedding vectors and define matrix $\mathbf{A}$ as follows:
\begin{equation}
a_{ij}=\frac{\mathbf{F}^{(l)}_{*i}\cdot\mathbf{F}^{(l)}_{*j}}{||\mathbf{F}^{(l)}_{*i}|| ||\mathbf{F}^{(l)}_{*j}||}
\label{j3_pre}
\end{equation}
$\mathbf{B}\in \{+1,-1\}^{c\times n}$ is the to-be-learned unified hash code matrix, and $c$ is the hash code length. In fact, Equation \eqref{j2} is a standard optimization problem for hashing learning. It is   NP-hard   to directly solve the equation \eqref{j2} due to the discrete limitation of $\mathbf{B}$. Thus we introduce an approximate hash function to get around this problem. Assuming that our hashing functions are linear functions of the learned deep feature of different modalities respectively. Our hash code approximating loss can be formulated as follows:
\begin{equation}
\mathcal{J}_3=\sum_{m=1}^2||\mathbf{B}-\mathbf{P}_i\mathbf{F}^{(m)}||_F^2
\label{j3_pre}
\end{equation}
where $\mathbf{P}_i$ is the linear hash function mapping deep feature $\mathbf{F}^{(m)}$ into common Hamming space. However, the learned common mapping hash function $\mathbf{P}_i$ only retains the modality specific properties, since the mapping function $\mathbf{P}_1$ and $\mathbf{P}_2$ are learned independently. To learn both inter-modal complementary relationships and intra-modal intrinsic relationships, we consider hash functions of different modalities with common component. Then our hash code approximating loss can be formulated as follows:
\begin{equation}
\mathcal{J}_3=\sum_{m=1}^2||\mathbf{B}-(\mathbf{C}+\mathbf{D}_m)\mathbf{F}^{(m)}||_F^2
\label{j3}
\end{equation}
where $\mathbf{C}$ is the common component of the mapping functions, $\mathbf{D}_1$ and $\mathbf{D}_2$ are individual components of the mapping functions. $\mathbf{C}$ is used to capture the common structure information while $\mathbf{D}_1$ and $\mathbf{D}_2$ are used to learn intra-modal intrinsic structures of repsective modalities.

To sum up, our whole object function is:
\begin{equation}
\begin{split}
\min_{\substack{\theta_x, \theta_y, \theta_l, \mathbf{B},\\
 \mathbf{C}, \mathbf{D}_1, \mathbf{D}_2}} \mathcal{J}=&-\sum_{m=1}^2\sum_{i,j=1}^{n}[s_{ij}\Phi_{ij}^m-log(1+\mathrm{e}^{\Phi_{ij}^m})]\\
&+\sum_{m=1}^2\alpha_m||\mathbf{B}-(\mathbf{C}+\mathbf{D}_m)\mathbf{F}^{(m)}||_F^2\\
&+\beta[||[(\mathbf{C}+\mathbf{D}_1)\mathbf{F}^{(1)}]^T(\mathbf{C}+\mathbf{D}_2)\mathbf{F}^{(2)}-\mathbf{A}||_F^2]\\
s.t.\quad \mathbf{B} \in & \{+1, -1\}^{c\times n}
\end{split}
\label{lossfunction}
\end{equation}
As we discussed above, the first term is the log likelihood loss term. By means of minimizing this term, we can preserve the  similarity relationships between the label embedding vectors and their deep features. The second term is hash code approximating loss and the loss for each modality is weighted by $\alpha_m$ We expect to learn hash functions while getting around of the hard discrete value matrix optimization problem. The third term is the hash code learning loss. By means of minimizing this term we can preserve the semantic attribute similarity. Overall, minimizing equation \eqref{lossfunction} makes two strings of hash code similar not only when the corresponding instances are of the same category but also when the instances are similar in semantic attributes.

\subsection{Optimization}
To address the problem (\ref{lossfunction}), we can not optimize all these parameters simultaneously. Thus, an alternative optimization algorithm is presented. We  learn one of the parameters at a time while fixing the others. We calculate the derivative of each parameter and use stochastic gradient descent (SGD) to update them. Note that,   we do not take $\mathbf{A}$ as parameter to be optimized, we only update $\mathbf{A}$ using Eq. \eqref{j3_pre} based on updated $\mathbf{F}_{*i}^{(l)}$ in each epoch and take it as a fixed supervised information.

\subsubsection{$\theta_x$($\theta_y$)-step}
To update the parameters of image network $\theta_x$, we first fix other parameters and calculate $\frac{\partial\mathcal{J}}{\partial\mathbf{F}^{(1)}_{*i}}$ for each training instance $\mathbf{x}_i$,  then calculate $\frac{\partial\mathcal{J}}{\partial\theta_x}$ via chain rule and update $\theta_x$ by BP algorithm. Thus, we calculate $\frac{\partial\mathcal{J}}{\partial\mathbf{F}^{(1)}_{*i}}$ and $\frac{\partial\mathcal{J}}{\partial\theta_x}$ as follows:
\begin{equation}\label{theta_x}
\begin{split}
  \frac{\partial\mathcal{J}}{\partial\mathbf{F}^{(1)}_{*i}}=&\frac12\sum_{j=1}^{n}(\sigma(\Phi^{{1}}_{ij})\mathbf{F}^{(l)}_{*j}-s_{ij}\mathbf{F}^{(l)}_{*j})\\
  &+2\alpha^1(\mathbf{C}+\mathbf{D}_i)^T[\mathbf{B}_{*i}-(\mathbf{C}+\mathbf{D}_1)\mathbf{F}_{*i}] \\
  &+2\beta\{\mathbf{P}_1^T\mathbf{P}_2\mathbf{F}^{(2)}[(\mathbf{P}_2\mathbf{F}^{(2)})^T\mathbf{P}_1\mathbf{F}^{(1)}_{*i}-(\mathbf{A}^T)_{*i}]\}
\end{split}
\end{equation}
\begin{equation}\label{chainrule1}
\begin{split}
  \frac{\partial\mathcal{J}}{\partial\theta_x}=\frac{\partial\mathcal{J}}{\partial\mathbf{F}^{(1)}_{*i}}\frac{\partial\mathbf{F}^{(1)}_{*i}}{\partial\theta_x}
\end{split}
\end{equation}
where $\mathbf{P}_1=\mathbf{C}+\mathbf{D}_1$, $\mathbf{P}_2=\mathbf{C}+\mathbf{D}_2$. And we can update $\theta_y$ in a similar way.

\subsubsection{$\theta_l$-step}
Just like we do with $\theta_x$ and $\theta_y$, to update the parameter of label transfer network $\theta_l$, we first calculate $\frac{\partial\mathcal{J}}{\partial\mathbf{F}^{(l)}_{*i}}$ and then calculate $\frac{\partial\mathcal{J}}{\partial\theta_l}$ using chain rule.
\begin{equation}\label{theta_l}
  \frac{\partial\mathcal{J}}{\partial\mathbf{F}^{(l)}_{*i}}=\frac12\sum_{m=1}^{2}\sum_{j=1}^{n}(\sigma(\Phi^{{m}}_{ij})\mathbf{F}^{(m)}_{*j}-s_{ij}\mathbf{F}^{(m)}_{*j})
\end{equation}
\begin{equation}\label{chainrule2}
\begin{split}
  \frac{\partial\mathcal{J}}{\partial\theta_l}=\frac{\partial\mathcal{J}}{\partial\mathbf{F}^{(l)}_{*i}}\frac{\partial\mathbf{F}^{(l)}_{*i}}{\partial\theta_l}
\end{split}
\end{equation}

\subsubsection{$\mathbf{D}_1$($\mathbf{D}_2$)-step}
The derivative of $\mathbf{D}_1$ is calculated as follows:
\begin{equation}\label{d}
\begin{split}
  \frac{\partial\mathcal{J}}{\partial\mathbf{D}_1}=&2\beta\mathbf{B}_2[\mathbf{B}_2^T(\mathbf{C}+\mathbf{D}_1F^{(1)})-\mathbf{A}^T(\mathbf{F}^{(1)})^T]\\
  &-2\alpha_1[\mathbf{B}-(\mathbf{C}+\mathbf{D}_1)\mathbf{F}^{(1)}](\mathbf{F}^{(1)})^T
\end{split}
\end{equation}
where $\mathbf{B}_2=(\mathbf{C}+\mathbf{D}_2)\mathbf{F}^{(2)}$, and we can update $\mathbf{D}_2$ in a similar way.

\subsubsection{$\mathbf{C}$-step}
The derivative of $\mathbf{C}$ is calculated as follows:
\begin{equation}\label{c}
\begin{split}
  \frac{\partial\mathcal{J}}{\partial\mathbf{C}}=&-2\sum_{m=1}^2\alpha_m[\mathbf{B}-(\mathbf{C}+\mathbf{D}_m)\mathbf{F}^{(m)}](\mathbf{F}^{(m)})^T\\
  &+2\beta\mathbf{P}_1\mathbf{F}^{(1)}[(\mathbf{P}_1\mathbf{F}^{(1)})^T\mathbf{P}_2\mathbf{F}^{(1)})-\mathbf{A}^T]\mathbf{F}^{(1)}\\
  &+2\beta\mathbf{P}_2\mathbf{F}^{(2)}[(\mathbf{P}_2\mathbf{F}^{(2)})^T\mathbf{P}_1\mathbf{F}^{(2)})-\mathbf{A}^T]\mathbf{F}^{(2)}
\end{split}
\end{equation}
where $\mathbf{P}_1=\mathbf{C}+\mathbf{D}_1$, $\mathbf{P}_2=\mathbf{C}+\mathbf{D}_2$.

\subsubsection{$\mathbf{B}$-step}
To update $\mathbf{B}$, the minimizing problem \eqref{lossfunction} is transformed into:
\begin{equation}
\begin{split}
\min_{\mathbf{B}} &\sum_{m=1}^2\alpha_m||\mathbf{B}-(\mathbf{C}+\mathbf{D}_m)\mathbf{F}^{(m)}||_F^2 \\
& s.t. \quad  \mathbf{B}\in \{-1,+1\}^{c\times n}
\end{split}
\end{equation}
we can easily solve this problem as follows:
\begin{equation}\label{b}
  \mathbf{B}=sgn[(\mathbf{C}+\mathbf{D}_1)\mathbf{F}^{(1)}+(\mathbf{C}+\mathbf{D}_2)\mathbf{F}^{(2)}]
\end{equation}
where $sgn(\bullet)$ is the signum function.

\section{Experiments}

\subsection{Experimental Setup}
We conduct experiments on two different benchmark datasets: NUS-WIDE \cite{chua2009nus-wide:} and AWA2  \cite{lampert2009learning}  to investigate the effectiveness of our LAEH.

The \emph{NUS-WIDE} dataset contains 269,648 images and the associated textual tags are collected from the Flickr website. Each instance is annotated with one or more labels from 81 concept labels. We take textual tags as the text modality and convert them into a series of 1000-dimensional BOW vectors. In this paper, we use instance pairs from 30 most frequent concepts. Note that the original \emph{NUS-WIDE} dataset can not directly  form disjoint classes between seen and unseen set as each instance pair has multiple labels. Thus, we only keep the most frequent label. As for zero-shot learning settings, we random select 10 categories of 30 as unseen categories and others as seen categories.

The \emph{AWA2} dataset consists of 30,475 images from 50 animal categories and 85 associated class-level attributes. It is a popular dataset for zero-shot learning. We also take textual tags as the text modality and convert into a series of 1000-dimensional BOW vectors. For zero-shot learning settings, we select 10 categories as unseen categories and the rest 40 categories as seen categories.

To verify the effectiveness of our method, seven related and representative methods are adopted for empirical comparison, including Deep Cross-Modal Hashing (DCMH) \cite{jiang2017deep}, Semantic Correlation Maximization (SCM-seq and SCM-orth) \cite{zhang2014large-scale}, Semantics Preserving Hashing (SePH) \cite{lin2015semantics}, Collective Matrix Factorization Hashing (CMFH) \cite{ding2014collective}, Attribute-Guided Network (AgNet) \cite{ji2020attribute-guided} and zero-shot cross-modal hashing (CZHash) \cite{liu2019cross}. In order to better study the advantages and disadvantages of compared methods, we take into account the diversity of comparison methods. SCM and SePH are traditional supervised cross-modal hashing methods. CMFH is an unsupervised CMH method, DCMH is a supervised CMH method based on deep neural networks. AgNet and CZHash are zero-shot CMH methods. For each compared method, we adopted or tuned the parameters based on the suggestions given in the shared codes or respective papers. For LAEH, we set $\alpha_1$=$\alpha_2$=1 and $\beta$=1.

We choose  the widely used evaluation metric MAP (mean average precision), to quantify the performance of hashing retrieval methods ~\cite{bronstein2010data,zhang2014large-scale},
\begin{equation}
MAP = \frac{1}{|\mathcal{Q}|}\sum_{i=1}^{|\mathcal{Q}|}\frac{1}{m_i}\sum_{j=1}^{m_i}P(j)
\end{equation}
where $\mathcal{Q}$ is the query set, $m_i$ is the number of ground-truth labeled relevant instances in the retrieval set for the $i$-th query. For each query, we obtain a ranking list of the retrieval set according to the Hamming distance between the query and each instance in the retrieval set. From the first instance in the rank list to the $j$-th relevant instance, we get the top $j$ retrieval subset and $P(j)$ is the precision of this subset. More intuitively, for each instance in query set $\mathcal{Q}$, MAP first calculates an average precision (AP) value and then takes the mean value of these APs. A larger MAP value indicates a better hashing retrieval performance.

\begin{table*}[]
\small
\centering
\scalebox{0.8}{
\begin{tabular}{c|l|llllll|llllll}
\hline
\multicolumn{2}{c|}{Dataset}        & \multicolumn{6}{c|}{NUS-wide}                                                                                                                                            & \multicolumn{6}{c}{AWA2}                                                                                                                                          \\ \hline
\multicolumn{2}{c|}{Class}          & \multicolumn{2}{c}{All classes}                           & \multicolumn{2}{c}{Unseen Classes}                              & \multicolumn{2}{c|}{Seen Classes}                               & \multicolumn{2}{c}{All classes}                           & \multicolumn{2}{c}{Unseen Classes}                              & \multicolumn{2}{c}{Seen Classes}                               \\ \hline
\multicolumn{2}{c|}{bits}           & \multicolumn{1}{c}{32bit} & \multicolumn{1}{c}{64bit} & \multicolumn{1}{c}{32bit} & \multicolumn{1}{c}{64bit} & \multicolumn{1}{c}{32bit} &
\multicolumn{1}{c|}{64bit} & \multicolumn{1}{c}{32bit} &
\multicolumn{1}{c}{64bit} & \multicolumn{1}{c}{32bit} &
\multicolumn{1}{c}{64bit} & \multicolumn{1}{c}{32bit} &
\multicolumn{1}{c}{64bit} \\ \hline
\multirow{8}{*}{I to T} & SCM-seq   &0.335 &0.344 &0.109 &0.108	&0.465	&0.467 &0.254 &0.259 &0.124	&0.128	&0.287	&0.292
\\
                        & SCM-orth  &0.325 &0.331 &0.104 &0.110	&0.440	&0.447 &0.238 &0.243 &0.119	&0.127	&0.268	&0.272
\\
                        & CMFH      &0.302 &0.306 &0.168 &0.164	&0.394	&0.398 &0.197 &0.208 &0.176	&0.178	&0.202	&0.216
\\
                        & SePH      &0.379 &0.382 &0.128 &0.132	&0.490	&0.497 &0.285 &0.296 &0.126	&0.134	&0.325	&0.337
\\
                        & DCMH      &0.457 &0.465 &0.124 &0.127	&\textbf{0.630}  &\textbf{0.641} &0.350 &0.356 &0.117 &0.119	&\textbf{0.416}	&\textbf{0.423}
 \\
                        & AgNet     &-&-&-&-&-&- &0.369 $\ast$	&0.374 $\ast$	&0.297	&0.301 &0.387 &0.392
 \\
                        & CZHash    &0.476 $\ast$ &0.482 $\ast$ &0.335 $\ast$ &0.346 $\ast$	&0.533	&0.546 &0.358 &0.361 &0.305 $\ast$ &0.306 $\ast$ &0.371 &0.375
 \\
                        & LAEH       &\textbf{0.558} &\textbf{0.564} &\textbf{0.340} &\textbf{0.349} &0.612 $\ast$	&0.618 $\ast$ &\textbf{0.374} &\textbf{0.381} &\textbf{0.327} &\textbf{0.328} &0.408 $\ast$ &0.417 $\ast$ \\ \hline
\multirow{8}{*}{T to I} & SCM-seq   &0.367 &0.368 &0.116 &0.121 &0.498	&0.497 &0.256 &0.261 &0.135	&0.139 &0.286 &0.292
\\
                        & SCM-orth  &0.368 &0.365 &0.125 &0.127	&0.495	&0.489 &0.243 &0.249 &0.113	&0.128 &0.276 &0.279
\\
                        & CMFH      &0.313 &0.315 &0.174 &0.176	&0.407	&0.409 &0.208 &0.210 &0.178	&0.175 &0.215 &0.219
\\
                        & SePH      &0.401 &0.407 &0.135 &0.136	&0.521	&0.529 &0.290 &0.301 &0.130	&0.135 &0.330 &0.342
\\
                        & DCMH      &0.481 &0.482 &0.129 &0.124	&\textbf{0.665}	&\textbf{0.668} &0.357 &0.362 &0.116	&0.118 &\textbf{0.425} &\textbf{0.431}
\\
                        & AgNet     &-&-&-&-&-&-  &0.380 $\ast$ &0.386 $\ast$	&0.296	&0.301 &0.401 &0.407
\\
                        & CZHash    &0.503 $\ast$ &0.510 $\ast$ &0.376 $\ast$ &0.379 $\ast$ &0.570 &0.572 &0.369 &0.365	&0.309 $\ast$	&0.313 $\ast$	&0.397	&0.391
 \\
                        & LAEH       &\textbf{0.585} &\textbf{0.596} &\textbf{0.387} &\textbf{0.391}	&0.635 $\ast$ &0.647 $\ast$ &\textbf{0.389} &\textbf{0.397}	&\textbf{0.343}	&\textbf{0.347}	&0.411 $\ast$	&0.419 $\ast$ \\ \hline
\end{tabular}}
\vspace{-0.5em}
\caption{Mean Average Precision (MAP) of all methods on AWA2 and NUS-wide. The best  and  second best are marked in bold and with *.}
\label{result}
\end{table*}

\begin{figure*}[bt]
\centering
\subfigure[MAP vs.  $\alpha_1$ and $\alpha_2$ (I to T)]{
\includegraphics[width=5cm]{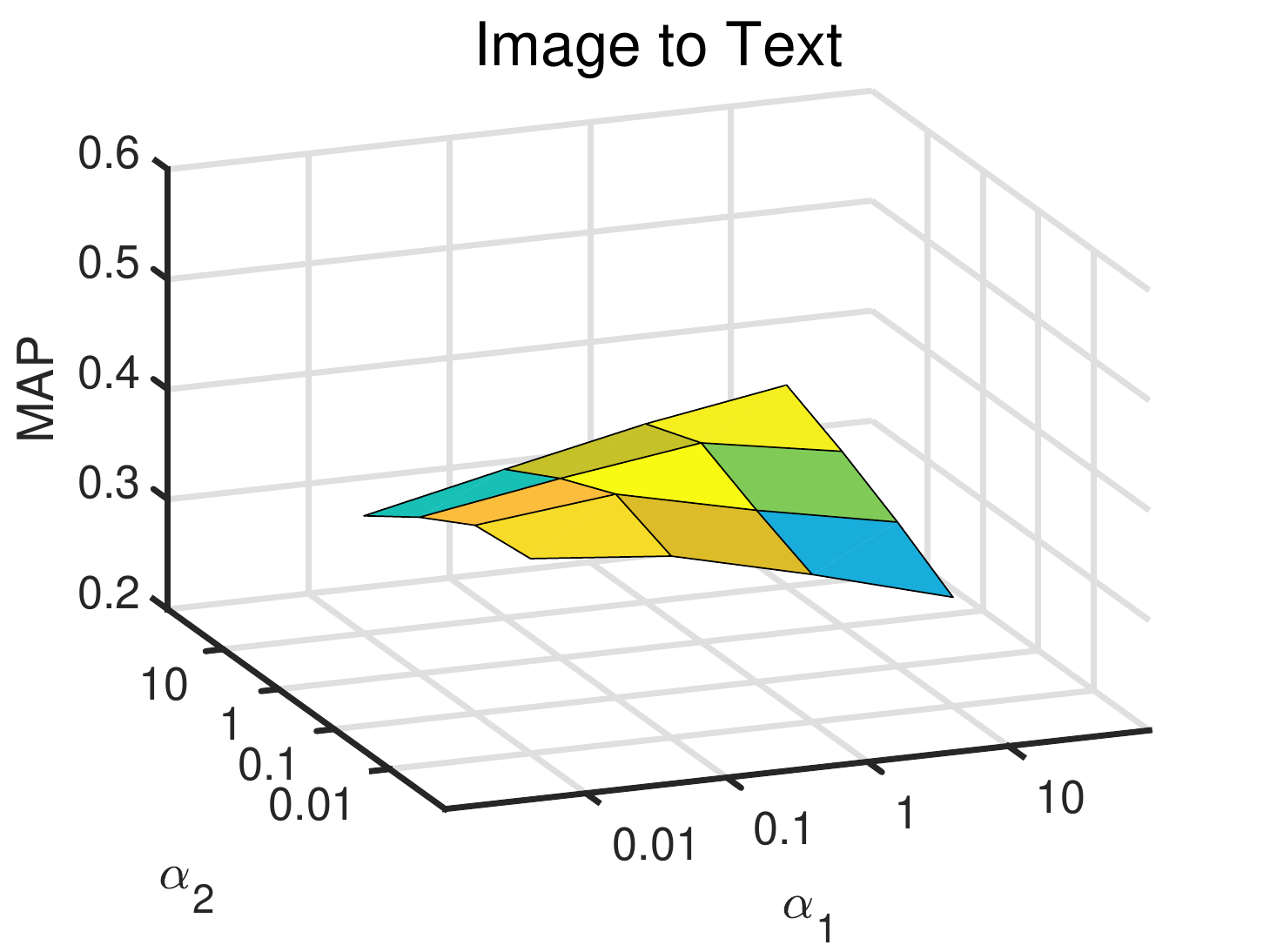}}
\subfigure[MAP vs.  $\alpha_1$ and $\alpha_2$ (T to I)]{
\includegraphics[width=5cm]{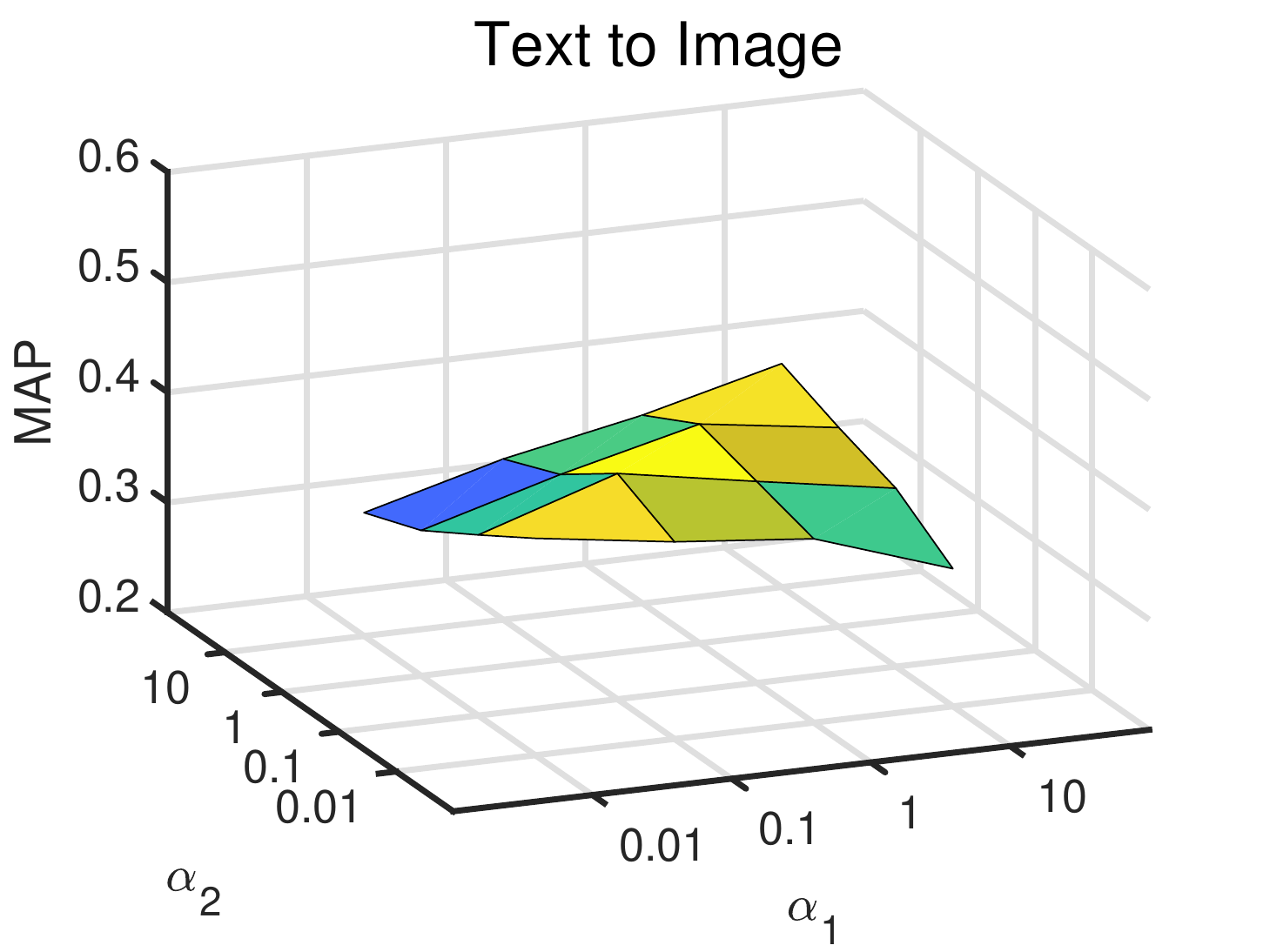}}
\subfigure[MAP vs.  $\beta$]{
\includegraphics[width=5cm]{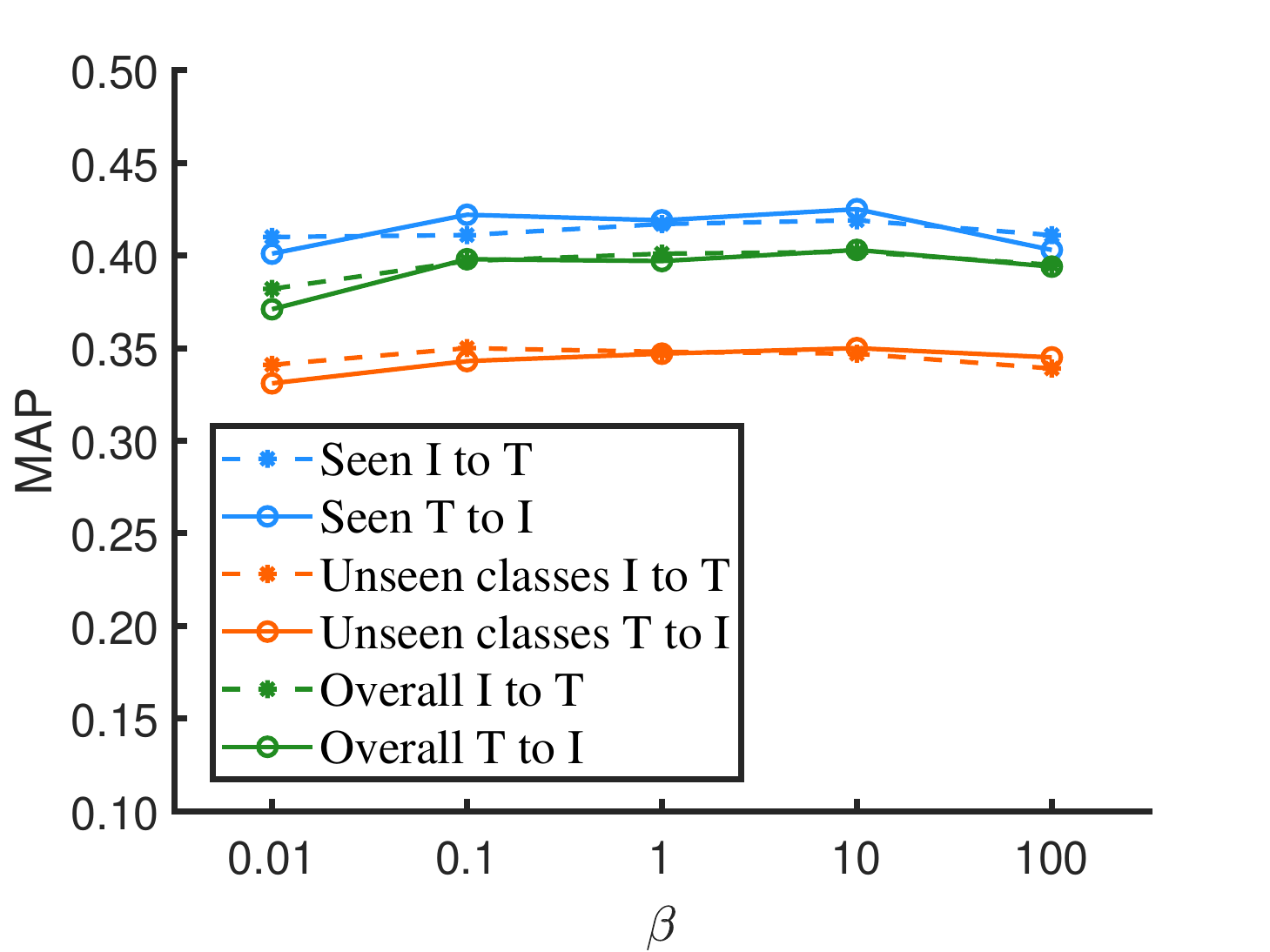}}
\vspace{-1em}
\caption{Performance of LAEH (MAP) under different settings of $\alpha_1$, $\alpha_2$ and $\beta$ on \emph{AWA2}.}
\label{alpha}
\end{figure*}

\subsection{Result and analysis}
We report the experimental results in Table \ref{result}, where ``I to T'' means the query is the image and the retrieval data is the text, and ``I to T'' means the query is the text and the retrieval data is the image.  The standard deviations of MAP results of compared methods are quite small (generally less than 3\%), and are thus omitted   in the table. Note that, AgNet requires the introduction of attribute vectors for each sample as supervised information which is not accessible for \emph{NUS-Wide} dateset. Thus, AgNet cannot be applied to \emph{NUS-Wide} dateset.
From Table \ref{result}, we have the following observations and conclusions.

Firstly, for query instances from seen classes, there are training samples from the same class, thus the problem degenerates to an ordinary cross-modal hashing problem. The emerging unseen class instances affect the search effectiveness of instances from seen classes to some extend. But the MAP on seen classes mainly reflects the effectiveness on ordinary cross-modal hashing problem of different methods. As we can see, supervised methods generally achieve better performance on seen classes than unsupervised method (CMFH), because supervised methods take advantage of class labels and aim to preserve the label similarity which makes hash codes similar among samples from the same class. Methods with deep architecture (DCMH, AgNet, CZHash, LAEH) generally achieve a better performance because of the stronger feature extraction capability than shallow ones.
CZHash considers both inter-modal and intra-modal similarity of hash codes, so the performance margin between CZHash and LAEH is smaller than that of other methods. LAEH performs a little worse than DCMH on seen classes because LAEH is guided by both kinds of supervised information (label similarity and attribute similarity), when the two types of guide information are not exactly consistent, the effectiveness will be worse than just using the label information when evaluating on seen classes.

Secondly, for samples from unseen classes, there are no training samples from the same class in the training set. For supervised CMH methods, the main idea is to preserve the similarity information in class label matrix. While unsupervised methods aim to discover and preserve the intrinsic structure of the cross-modal data. Therefore unsupervised methods can learn a part of intrinsic structure of the cross-modal data. For example, samples with similar semantic attributes will have similar features and unsupervised methods mainly aim to preserve this features similarity. Thus, unsupervised method CMFH achieves a better performance on unseen classes than supervised CMH methods. AgNet, CZHash and LAEH are designed for zero-shot scenarios, thus they can transfer attribute knowledge from seen classes to unseen ones and achieve a better performance than CMFH on unseen classes. LAEH achieves a better performances than other zero-shot CMH methods on unseen classes. That is because LAEH uses attribute similarity as a supplement of label similarity and further refines the label attributes by considering multi-modal training data. This makes the learned hash codes similar only when the two samples are in the same class and with similar semantic attributes.

Thirdly, we can observe that zero-shot learning methods, as well as LAEH, generally achieve better performances on \emph{NUS-wide} dataset than on \emph{AWA2} dataset. On the one hand, it is because   in our experimental settings, there are more classes in \emph{AWA2} dataset,   making the hashing problem more challenging. On the other hand, \emph{AWA2} dataset contains only animal-related samples while the inter-class correlation in the \emph{NUS-wide} dataset is relatively low. Zero-shot CMH methods mainly rely on semantic attribute knowledge learned from seen classes, the stronger the diversity of unseen classes is, the higher the distinction of class attributes is and the more effective  zero-shot CMH methods will be. For LAEH, the semantic embedding vectors are initialized from a pre-trained word2vec model, which learns from plenty text from \emph{wikipedia}. The lower the correlation between classes, the better the effect of attribute (semantic) embedding. Thus, LEA works better on datasets with high class diversity.

\subsection{Parameter sensitivity analysis}
In this part, we evaluate the sensitivity of the hyper-parameters $\alpha_1$, $\alpha_2$ and $\beta$, which control the   weights of the different components of the loss function.
For parameter $\alpha_1$ and $\alpha_2$, we conduct experiments on \emph{AWA2} dataset with $\beta=1$ and $c=64$. We can see from Figure \ref{alpha} that, when $\alpha_1=\alpha_2$, LAEH is not sensitive to the exact numerical size of $\alpha_1$ and $\alpha_2$. However,  when $\alpha_1\neq\alpha_2$, the bigger the gap between $\alpha_1$ and $\alpha_2$, the lower the MAP value of LAEH is. That is because   if the gap between $\alpha_1$ and $\alpha_2$ is large, the loss on one modality is discounted,   compromising the retrieval performance.

For parameter $\beta$, we run experiments on \emph{AWA2} dataset with $\alpha_1$=$\alpha_2$=1, $c$=64. From Fig. \ref{alpha}, we   observe that, LAEH is not sensitive to $\beta$ when $0.1\leq\beta\leq10$. When $\beta$ is too large ($\beta$=100) or too small ($\beta$=0.01), the MAP of LAEH   slightly decreases. This is because that we adopt an alternative optimization algorithm and $\beta$ only affects the optimization of part of the parameters. Adequate iteration further reduces the influence of $\beta$. Thus, only when $\beta$ is too large or too small,  it does influence the   effectiveness of LAEH. From these observations, we adopt $\alpha_1$=$\alpha_2$=1, $\beta$=1 in the experiments.

\section{Conclusion}
In this paper, we proposed a novel method called Label Attribute Embedding Hashing (LAEH) for zero-shot cross-modal retrieval. LAEH uses a pre-trained Word2Vec model to get the semantic embedding of each sample class and introduces a transformer network to make the semantic embeddings more compatible with the hash learning architecture.
Experiments on benchmark datasets show that LAEH can achieve the state-of-the-art performance in zero-shot cross-modal hashing problems. In the future work, we will study the class imbalance problem in zero/few-shot CMH.

\bibliographystyle{named}
\bibliography{lae}

\end{document}